# Structure-Based Causes and Explanations in the Independent Choice Logic

Alberto Finzi and Thomas Lukasiewicz*
Dipartimento di Informatica e Sistemistica,
Università di Roma "La Sapienza"
Via Salaria 113, I-00198 Rome, Italy
{finzi, lukasiewicz}@dis.uniroma1.it

## Abstract

This paper is directed towards combining Pearl's structural-model approach to causal reasoning with high-level formalisms for reasoning about actions. More precisely, we present a combination of Pearl's structural-model approach with Poole's independent choice logic. We show how probabilistic theories in the independent choice logic can be mapped to probabilistic causal models. This mapping provides the independent choice logic with appealing concepts of causality and explanation from the structural-model approach. We illustrate this along Halpern and Pearl's sophisticated notions of actual cause, explanation, and partial explanation. This mapping also adds first-order modeling capabilities and explicit actions to the structural-model approach.

## 1 INTRODUCTION

Handling causality is an important issue, which emerges in many applications in AI. The existing approaches to causality in the AI literature can be roughly divided into those that have been developed as modal nonmonotonic logics (especially in the context of logic programming) and those that evolved from the area of Bayesian networks. A representative of the former is Geffner's modal nonmonotonic logic for handling causal knowledge [14, 15], which has been inspired by default reasoning from conditional knowledge bases. More specialized modal-logic based formalisms play an important role in dealing with causal knowledge about actions and change; see especially the work by Turner [36] and the references therein for an overview. A representative of the latter is Pearl's approach to modeling causality by structural equations [2, 12, 28, 29], which is central to a number of recent research efforts. In particular, the evaluation of deterministic and probabilistic counterfactuals has been explored, which is at the core of problems in fault diagnosis, planning, decision making, and determination of liability [2]. It has been shown that the structural-model approach allows a precise modeling of many important causal relationships, which can especially be used in natural language processing [12]. An axiomatization of reasoning about causal formulas in the structural-model approach has been given by Halpern [16].

Concepts of causality also play an important role in the generation of explanations, which are of crucial importance in areas like planning, diagnosis, natural language processing, and probabilistic inference. Different notions of explanations have been studied quite extensively, see especially [19, 13, 34] for philosophical work, and [27, 35, 20] for work in AI that is related to Bayesian networks. A critical examination of such approaches from the viewpoint of explanations in probabilistic systems is given in [6].

In recent papers [17, 18], Halpern and Pearl formalized causality using a model-based definition, which allows for a precise modeling of many important causal relationships. Using a notion of weak cause, they propose appealing definitions of actual cause [17] and of causal explanation [18]. As they show, their notions of actual cause and causal explanation, which is very different from the concept of causal explanation in [24, 26, 14], models well many problematic examples in the literature. As for computation, Eiter and Lukasiewicz [7, 8, 9] analyzed the complexity of these notions and identified tractable cases, and Hopkins [21] explored search-based strategies for computing actual causes in the general and restricted settings.

However, structural causal models, and thus also the above notions of actual cause and causal explanation, have only a limited expressiveness in the sense that (i) they do not allow for first-order modeling, and (ii) they only allow for explicitly setting the values of endogenous variables (also called an *intervention*) as actions, but not for explicit actions as in well-known formalisms for reasoning about actions.

There are a number of formalisms for probabilistic reason-

---

ing about actions. In particular, Bacchus et al. [1] propose a probabilistic generalization of the situation calculus, which is based on first-order logics of probability, and which allows one to reason about an agent's probabilistic degrees of belief and how these beliefs change when actions are executed. Poole's independent choice logic [30, 31] is based on acyclic logic programs under different "choices". Each choice along with the acyclic logic program produces a first-order model. By placing a probability distribution over the different choices, one then obtains a distribution over the set of first-order models. Other probabilistic extensions of the situation calculus are given in [25, 11]. A probabilistic extension of the action language $\mathcal{A}$ is given in [3].

The main idea behind this paper is to develop a combination of Pearl's structural-model approach to (probabilistic) causal reasoning with high-level formalisms for (probabilistic) reasoning about actions. To this end, we present a combination of Pearl's structural-model approach with Poole's independent choice logic. We show how probabilistic theories in the independent choice logic [30, 31] can be translated into probabilistic causal models. This translation provides the independent choice logic with appealing concepts of causality and explanation from the structural-model approach. We illustrate this along Halpern and Pearl's notions of actual cause and causal explanation. This mapping also adds first-order modeling capabilities and explicit actions to the structural-model approach.

The work closest in spirit to this paper is perhaps the recent one by Hopkins and Pearl [22], which combines the situation calculus [33] with the structural model-approach. However, the generated causal models are much different from the ones in this paper. First, and as a central conceptual difference, Hopkins and Pearl consider a standard situation calculus formalization, which allows for expressing uncertainty about the initial situation, but which does not allow for uncertain effects of actions. In this paper, however, we consider Poole's independent choice logic [30, 31], which is a first-order formalism that allows both for probabilistic uncertainty about the initial situation and about the effects of actions. Second, the work [22] focuses only on counterfactual and probabilistic counterfactual reasoning, while our work here extends the notions of actual cause, explanation, and partial explanation to the independent choice logic. Third, [22] focuses only on hypothetical reasoning about subsequences of an initially fixed sequence of actions, while our approach here basically allows for hypothetical reasoning about any actions and fluent values.

Note that also Poole [32] defines a notion of explanation for his independent choice logic. However, Poole's notion of explanation in [32] is based on abductive reasoning and assumes that explanations are defined over choice atoms. Our notion of explanation in this paper, in contrast, is based on causal reasoning in structural causal models, and assumes that explanations are defined over endogenous variables. Hence, our concept of explanation here is conceptually much different from the one by Poole in [32].

## 2 CAUSAL MODELS

In this section, we recall some basic concepts from Pearl's structural-model approach to causality [2, 12, 28, 29]. In particular, we recall causal and probabilistic causal models.

### 2.1 PRELIMINARIES

We assume a set of *random variables*. Every variable $X_i$ may take on *values* from a nonempty *domain* $D(X_i)$. A *value* for a set of variables $X$ is a mapping $x$ that associates with each $X_i \in X$ an element of $D(X_i)$ (for $X = \emptyset$, the unique value is the empty mapping $\emptyset$). The *domain* of $X$, denoted $D(X)$, is the set of all values for $X$. For $Y \subseteq X$ and $x \in D(X)$, denote by $x|Y$ the restriction of $x$ to $Y$. For disjoint sets of variables $X, Y$, and values $x \in D(X)$, $y \in D(Y)$, denote by $xy$ the union of $x$ and $y$. We often identify singletons $\{X_i\}$ with $X_i$, and their values $x$ with $x(X_i)$.

### 2.2 CAUSAL MODELS

A *causal model* $M = (U, V, F)$ consists of two disjoint sets $U$ and $V$ of *exogenous* and *endogenous* variables, respectively, and a set $F = \{F_X \mid X \in V\}$ of functions $F_X : D(PA_X) \to D(X)$ that assign a value of $X$ to each value of the *parents* $PA_X \subseteq U \cup V \setminus \{X\}$ of $X$. The values $u \in D(U)$ are also called *contexts*. A *probabilistic causal model* $M = ((U, V, F), P)$ consists of a causal model $(U, V, F)$ and a probability function $P$ on $D(U)$.

We focus here on the principal class [17] of *recursive* causal models $M = (U, V, F)$ in which a total ordering $\prec$ on $V$ exists such that $Y \in PA_X$ implies $Y \prec X$, for all $X, Y \in V$. In such models, every assignment to the exogenous variables $U = u$ determines a unique value $y$ for every set of endogenous variables $Y \subseteq V$, denoted $Y_M(u)$ (or simply $Y(u)$). For any causal model $M = (U, V, F)$, set of variables $X \subseteq V$, and value $x \in D(X)$, the causal model $M_{X=x} = (U, V \setminus X, F_{X=x})$, where $F_{X=x} = \{F'_Y \mid Y \in V \setminus X\}$ and each $F'_Y$ is obtained from $F_Y$ by setting $X$ to $x$, is a *submodel* of $M$. We abbreviate $M_{X=x}$ and $F_{X=x}$ by $M_x$ and $F_x$, respectively. For $Y \subseteq V$, we abbreviate $Y_{M_x}(u)$ by $Y_x(u)$. A causal or probabilistic causal model is *binary* iff $|D(X)| = 2$ for all endogenous variables $X$.

**Example 2.1** *(stopping robot)* Suppose a mobile robot detects the presence of an obstacle. Then, the command stop is executed by the control system, which activates two brakes 1 and 2 (wheels behind and ahead), and the robot stops. The robot can stop using only one of the two brakes.

This scenario can be modeled by the following recursive causal model $M = (U, V, F)$. The exogenous variables are given by $U = \{U_s\}$, where $D(U_s) = \{0, 1\}$, and $U_s$ is 1 iff an obstacle has been detected. The endogenous variables are given by $V = \{CS, B_1, B_2, S\}$, where $D(X) = \{0, 1\}$ for all $X \in V$, $CS$ is 1 iff the command stop is executed, $B_i$ is 1 iff brake $i$ is activated, and $S$ is 1 iff the robot stops. The functions $F = \{F_X \mid X \in V\}$ are given by $F_{CS} = U_s$, $F_{B_1} = F_{B_2} = CS$, and $F_S = 1$ iff $B_1 = 1 \vee B_2 = 1$. Fig. 1 shows the parent relationships between the variables.

The submodel $M_{B_2=0} = (U, V_{B_2=0}, F_{B_2=0})$ is given by $V_{B_2=0} = \{CS, B_1, S\}$ and $F_{B_2=0} = \{F'_{CS} = F_{CS}, F'_{B_1} = F_{B_1}, F'_S = 1 \text{ iff } B_1 = 1\}$. Then, $S_{B_2=0}(U_s = 1) = 1$.

A probabilistic causal model $(M, P)$ may then be given by the additional probability function $P$ on $D(U)$ defined by $P(U_s = 1) = 0.7$ and $P(U_s = 0) = 0.3$. □

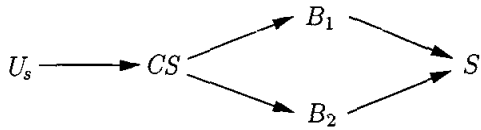

Figure 1: Causal Graph

## 3 INDEPENDENT CHOICE LOGIC

In this section, we recall Poole's independent choice logic (ICL) [30, 31, 32]. We first recall a many-sorted first-order language of logic programs, which are given a semantics in Herbrand interpretations. We then recall the ICL itself.

### 3.1 PRELIMINARIES

Let $\Phi$ be a many-sorted first-order vocabulary with the sorts *object*, *time*, and *action*. Let $\Phi$ contain function symbols of the sort $object^k \to object$, the function symbols 0 and +1 of the sorts *time* and $time \to time$, respectively, and function symbols of the sort $object^k \to action$, where $k \geq 0$. We call them *object*, *time*, and *action symbols*, respectively. As usual, *constant symbols* are 0-ary function symbols. Let $\Phi$ also contain predicate symbols of the sort $object^k \times time$, where $k \geq 0$, and the predicate symbol $do$ of the sort $action \times time$. We call them *fluent* and *action predicate symbols*, respectively. Let $X$ be a set of variables, which are divided into *object*, *time*, and *action* variables.

An *object term* is either an object variable from $X$ or an expression of the form $f(t_1, \ldots, t_k)$, where $f$ is a $k$-ary object symbol, and $t_1, \ldots, t_k$ are object terms. A *time term* is either a time variable from $X$, or the time symbol 0, or an expression of the form $s+1$, where $s$ is a time term. We use $1, 2, \ldots$ to abbreviate $0+1, 0+1+1, \ldots$. An *action term* is either an action variable from $X$, or an expression of the form $a(t_1, \ldots, t_k)$, where $a$ is a $k$-ary action symbol, and $t_1, \ldots, t_k$ are object terms.

We define *formulas* by induction as follows. The propositional constants *false* and *true*, denoted $\bot$ and $\top$, respectively, are formulas. *Atomic* formulas (or *atoms*) are of the form $do(a, s)$ or $p(t_1, \ldots, t_k, s)$, where $a$ is an action term, $s$ is a time term, $p$ is a $k$-ary fluent predicate symbol, and $t_1, \ldots, t_k$ are object terms. We call them *action atoms* and *fluent atoms* (or simply *actions* and *fluents*), respectively. If $\phi$ and $\psi$ are formulas, then also $\neg \phi$ and $(\phi \wedge \psi)$. We use $(\phi \vee \psi)$ and $(\phi \Leftarrow \psi)$ to abbreviate $\neg(\neg \phi \wedge \neg \psi)$ and $\neg(\neg \phi \wedge \psi)$, respectively, and adopt the usual conventions to eliminate parentheses. A *clause* is a formula of the form $\phi \Leftarrow \psi$, where $\phi$ (resp., $\psi$) is an atom (resp., formula) called its *head* (resp., *body*).

Terms and formulas are *ground* iff they do not contain any variables. Substitutions, ground substitutions, and ground instances of terms and formulas are defined as usual.

We use $HB_\Phi$ (resp., $HU_\Phi$) to denote the Herbrand base (resp., Herbrand universe) over $\Phi$. A *world* $I$ is a subset of $HB_\Phi$. We use $\mathcal{I}_\Phi$ to denote the set of all worlds over $\Phi$. A *variable assignment* $\sigma$ maps every variable from $X$ to an element of $HU_\Phi$ of appropriate sort. It is extended to object, time, and action terms as usual. The *truth* of formulas $\phi$ in $I$ under $\sigma$, denoted $I \models_\sigma \phi$, is defined by induction as follows (we write $I \models \phi$ when $\phi$ is ground):

- $I \models_\sigma do(a, s)$ iff $do(\sigma(a), \sigma(s)) \in I$;
- $I \models_\sigma p(t_1, \ldots, t_k)$ iff $p(\sigma(t_1), \ldots, \sigma(t_k)) \in I$;
- $I \models_\sigma \neg \phi$ iff not $I \models_\sigma \phi$;
- $I \models_\sigma (\phi \wedge \psi)$ iff $I \models_\sigma \phi$ and $I \models_\sigma \psi$.

A world $I$ is a *model* of a set of formulas $\mathcal{F}$, denoted $I \models \mathcal{F}$, iff $I \models_\sigma F$ for all $F \in \mathcal{F}$ and all $\sigma$.

### 3.2 INDEPENDENT CHOICE LOGIC

A *choice space* $C$ is a set of pairwise disjoint and nonempty sets $A \subseteq HB_\Phi$. The members of $C$ are called its *alternatives* and their elements *atomic choices*. A *total choice* of $C$ is a set $B \subseteq HB_\Phi$ such that $|B \cap A| = 1$ for all $A \in C$. A *probability* $P$ on a choice space $C$ is a probability function on the set of all total choices of $C$. If $C$ and all its alternatives are finite, then $P$ can be defined by (i) a mapping $P \colon \bigcup C \to [0, 1]$ such that $\sum_{a \in A} P(a) = 1$ for all $A \in C$, and (ii) $P(B) = \Pi_{b \in B} P(b)$ for all total choices $B$ of $C$.

A *logic program* $L$ is a set of clauses. We use $ground(L)$ to denote the set of all ground instances of clauses in $L$. A logic program is *acyclic* iff a mapping $\kappa$ from $HB_\Phi$ to the non-negative integers exists such that $\kappa(p) > \kappa(q)$ for all $p, q \in HB_\Phi$ where $p$ (resp., $q$) occurs in the head (resp., body) of some clause in $ground(L)$. The *answer set* (or *stable model*) of an acyclic logic program $L$ is a world $I$ such that for every $p \in HB_\Phi$, it holds that $I \models p$ iff $I \models \psi$ for some clause $p \Leftarrow \psi$ in $ground(L)$.

An *independent choice logic theory* (or *ICL-theory*) $T = (C, L)$ consists of a choice space $C$, and an acyclic logic program $L$ such that no atomic choice in $C$ coincides with the head of any clause in $ground(L)$. A *probabilistic ICL-theory* (or *PICL-theory*) $T = ((C, L), P)$ consists of an ICL-theory $(C, L)$ and a probability $P$ on $C$.

We next define the semantics of ICL- and PICL-theories by associating with them certain worlds and a probability distribution on certain worlds, respectively. A world $I$ is a *model* of an ICL-theory $T = (C, L)$, denoted $I \models T$, iff $I$ is an answer set of $L \cup \{p \Leftarrow \top \mid p \in B\}$ for some total choice $B$ of $C$. In a PICL-theory $T = ((C, L), P)$, the *probability* of such a world $I$ is then defined as $P(B)$.

The following example illustrates how action descriptions and probabilistic action descriptions can be encoded in ICL- and PICL-theories, respectively.

**Example 3.1** *(mobile robot)* Consider a mobile robot, which can navigate in an environment and pick up objects. We assume the constants $r_1$ (robot), $o_1$ (object), $p_1, p_2$ (positions), and 0 (time). The domain is described by the fluents $carrying(O, T)$ and $at(X, Pos, T)$, where $O \in \{o_1\}$, $X \in \{r_1, o_1\}$, $Pos \in \{p_1, p_2\}$, and $T \in \{0, 1, \ldots\}$. Here, $carrying(o, t)$ and $at(x, p, t)$ mean that the robot $r_1$ is carrying the object $o$ at time point $t$ and that the robot or object $x$ is at position $p$ at time point $t$, respectively. The robot is endowed with the actions $moveTo(Pos)$, $pickUp(O)$, and $putDown(O)$, where $Pos \in \{p_1, p_2\}$ and $O \in \{o_1\}$. Here, $moveTo(p)$, $pickUp(o)$, and $putDown(o)$ represent the actions "move to the position $p$", "pick up the object $o$", and "put down the object $o$", respectively. The action $pickUp(o)$ is stochastic: It is not reliable, and thus can fail. Furthermore, we have the predicates $do(A, T)$, which represents the execution of an action $A$ at time $T$, and $fa(A, T)$ (resp., $su(A, T)$), which represents the failure (resp., success) of an action $A$ executed at time $T$. An ICL-theory $(C, L)$ is then given by the choice space $C = \{\{fa_t = fa(pickUp(o_1), t), su_t = su(pickUp(o_1), t)\} \mid t \in \{0, 1\}\}$ and the following logic program $L$:

$carrying(O, T+1) \Leftarrow at(r_1, Pos, T) \land at(O, Pos, T) \land$
$\quad do(pickUp(O), T) \land su(pickUp(O), T);$
$carrying(O, T+1) \Leftarrow carrying(O, T) \land$
$\quad \neg do(putDown(O), T);$
$at(r_1, Pos, T+1) \Leftarrow do(moveTo(Pos), T);$
$at(r_1, Pos, T+1) \Leftarrow at(r_1, Pos, T) \land$
$\quad \neg do(moveTo(Pos1), T) \land Pos1 \neq Pos;$
$at(O, Pos, T+1) \Leftarrow at(O, Pos, T) \land \neg carrying(O, T);$
$at(O, Pos, T+1) \Leftarrow carrying(O, T) \land$
$\quad do(moveTo(Pos), T);$
$at(O, Pos, T+1) \Leftarrow at(O, Pos, T) \land$
$\quad \neg do(moveTo(Pos1), T) \land Pos1 \neq Pos;$
$at(o_1, p_2, 0) \Leftarrow \top; \ at(r_1, p_2, 0) \Leftarrow \top.$

A PICL-theory $((C, L), P)$ is given by $P(\{fa_0, su_1\}) = P(\{su_0, fa_1\}) = 0.21$, $P(\{fa_0, fa_1\}) = 0.09$, and $P(\{su_0, su_1\}) = 0.49$, which is obtained from $P(fa_0) = P(fa_1) = 0.3$ and $P(su_0) = P(su_1) = 0.7$ by assuming probabilistic independence between $\{fa_0, su_0\}$ and $\{fa_1, su_1\}$. □

## 4 TRANSLATIONS

In this section, we first give a translation of PICL-theories into probabilistic causal models. We then show how action executions at different time points can be included into this translation. We finally also provide a converse translation of binary probabilistic causal models into PICL-theories.

### 4.1 FROM ICL TO CAUSAL MODELS

We now define a translation of PICL-theories $((C, L), P)$ into probabilistic causal models. The main idea behind it is to use (i) each alternative $A$ of the choice space $C$ as an exogenous variable with the set of all atomic choices of $A$ as domain, and (ii) each other ground atom as an endogenous variable with binary domain, where the functions are specified by the clauses of the acyclic logic program $L$.

In the sequel, let $T = ((C, L), P)$ be a PICL-theory. The probabilistic causal model associated with $T$, denoted $M_T = ((U_T, V_T, F_T), P_T)$, is defined as follows:

- $U_T = C$, where $D(A) = A$ for all $A \in C$.

- $V_T = HB_\Phi \setminus \bigcup C$, where $D(V_i) = \{\bot, \top\}$ for all $V_i \in V_T$.

- $F_T = \{F_p \mid p \in V_T\}$, where $PA_p$ is the set of all ground atoms that occur in the body of some $p \Leftarrow \psi$ in $ground(L)$, and for every $v \in D(PA_p)$ we define $F_p(v) = \top$ iff $v \models \psi$ for some $p \Leftarrow \psi$ in $ground(L)$. Notice that then $F_p = \bot$ for all ground atoms $p$ in no head of a clause in $ground(L)$.

- $P_T(u) = P(\{u(A) \mid A \in C\})$ for all $u \in D(U)$.

For ICL-theories $T = (C, L)$, the causal model associated with $T$, denoted $M_T$, is defined as the above $(U_T, V_T, F_T)$. Given a total choice $B$ for $C$, we define $u_B \in D(U_T)$ by $u_B(A) \in B \cap A$ for all $A \in U_T = C$. The following example illustrates the above translation of PICL-theories into probabilistic causal models.

**Example 4.1** *(mobile robot (continued))* Consider the PICL-theory $T = ((C, L), P)$ given in Example 3.1. Its associated probabilistic causal model $M_T = ((U_T, V_T, F_T), P_T)$ is given as follows. The exogenous variables are given by $U_T = \{U_0, U_1\}$, where $U_0 = \{fa_0, su_0\} = D(U_0)$ and $U_1 = \{fa_1, su_1\} = D(U_1)$. The endogenous variables $V_T$ are given by all the ground atoms that do not occur in $U_T$, where $D(X) = \{\bot, \top\}$ for all $X \in V_T$. For example, the ground atoms $carrying(o_1, 0)$, $at(r_1, p_1, 1)$, $do(moveTo(p_1), 0)$, and $do(pickUp(o_1), 1)$ are all in $V_T$. The functions $F_T = \{F_p \mid p \in V\}$ are as specified above. For example, $F_{at(r_1, p_1, 1)} = \top$ iff either $do(moveTo(p_1), 0) = \top$, or $at(r_1, p_1, 0) = \top$ and $do(moveTo(p_2), 0) = \bot$. Finally, $P_T$ is defined by $P_T(u) = P(B_u)$ for all $u \in D(U)$, where $B_u$ is the total choice associated with $u$. For example, $P_T(U_0 = fa_0, U_1 = fa_1) = P(\{fa_0, fa_1\}) = 0.09$. □

### 4.2 ACTION EXECUTION SETS

We next describe how action executions at different time points in ICL can be incorporated into the probabilistic causal model $M_T$ associated with a PICL-theory $T$.

We define an *action execution set* $E$ as a set of ground atoms of the form $do(\alpha, t)$. Intuitively, $E$ represents the following set of action executions: For every $do(\alpha, t) \in E$, the action $\alpha$ is executed at time point $t$.

**Example 4.2** *(mobile robot (continued))* An action execution set for the PICL-theory of Example 3.1 is given by $E = \{do(moveTo(p_1), 0), do(pickUp(o_1), 1)\}$, which represents the execution of the actions "move to $p_1$" and "pick up $o_1$" at time points 0 and 1, respectively. □

In the sequel, let $T = ((C, L), P)$ be a PICL-theory. An action execution set $E$ can be taken into consideration in $M_T$ either (1) by additionally expressing the elements of $E$ as clauses in $L$ and then generating $M_T$, or (2) by considering the submodel of $M_T$ in which the elements of $E$ are explicitly set to $\top$. More formally, the probabilistic causal model $M = ((U, V, F), P)$ for $T$ and $E$ are defined as follows:

(1) We define $M$ as $M_{T'}$, where $T' = ((C, L \cup \{e \Leftarrow \top \mid e \in E\}), P)$. Then, the executions in $E$ can be overridden by explicitly setting them in $M$ (for example, in counterfactual reasoning).

(2) We define $M$ as $(M_T)_E = (M_T)_{E=e}$, where $e \in D(E)$ is defined by $e(X) = \top$ for all $X \in E$. Then, the executions in $E$ are fixed in $M$: They do not occur in $M$ and thus cannot be changed anymore.

Since from a technical viewpoint, (1) is a special case of (2), we consider only (2) in the rest of this paper. The following example illustrates (1) and (2).

**Example 4.3** *(mobile robot (continued))* Consider the PICL-theory $T = ((C, L), P)$ given in Example 3.1, and the action execution set $E$ given in Example 4.2. Under the representation (1), we then obtain the probabilistic causal model $M_{T'}$, where $T' = ((C, L'), P)$ and $L' = L \cup \{do(moveTo(p_1), 0) \Leftarrow \top, do(pickUp(o_1), 1) \Leftarrow \top\}$. Here, alternative executions can be explored by considering submodels that are obtained from $M_{T'}$ by explicitly setting the values of $do(moveTo(p_1), 0)$ and $do(pickUp(o_1), 1)$, for example, to $\bot$ and $\top$, respectively. Under (2), we obtain the causal model $(M_T)_E = (M_T)_{E=e}$, where $e(X) = \top$ for all $X \in E$. Here, the action executions in $E$ are fixed and cannot be changed anymore. □

### 4.3 FROM CAUSAL MODELS TO ICL

We finally also provide a converse translation of binary probabilistic causal models $M = ((U, V, F), P)$ into PICL-theories. The main ideas behind it are (i) to use the domains of the exogenous variables in $U$ as alternatives in the choice space, and (ii) to represent the functions in $F$ as clauses in an acyclic logic program under the answer set semantics.

In the sequel, let $M = ((U, V, F), P)$ be a binary probabilistic causal model, where $D(X) = \{0, 1\}$ for all $X \in V$, $F = \{F_X \mid X \in V\}$, and $PA_X$ is finite for every $X \in V$. The PICL-theory associated with $M$, denoted $T_M = ((C_M, L_M), P_M)$, is then defined as follows:

- $C_M = \{\{U_i = u_i \mid u_i \in D(U_i)\} \mid U_i \in U\}$;
- $L_M = \{X = 1 \Leftarrow \phi_i \mid X \in V, i \in I\}$, where $\phi_i$, for every index $i \in I$, is a conjunction of assignments $Y = y$, such that (i) $Y \in PA_X$, (ii) $y \in D(Y)$, and (iii) for all $p \in D(PA_X)$, it holds that $\bigvee_{i \in I} \phi_i$ is true in $p$ iff $F_X(p) = 1$;
- $P_M(B) = P(\{(U_i, u_i) \mid U_i = u_i \in B\})$ for every total choice $B$ of $C_M$.

For binary causal models $M = (U, V, F)$, the ICL-theory associated with $M$, denoted $T_M$, is defined as the above $(C_M, L_M)$. The following example illustrates the translation of probabilistic causal models into PICL-theories.

**Example 4.4** *(stopping robot (continued))* Consider again the probabilistic causal model $M = ((U, V, F), P)$ given in Example 2.1. Its associated PICL-theory $T_M = ((C_M, L_M), P_M)$ is given as follows. The choice space is given by $C_M = \{\{U_s = 1, U_s = 0\}\}$. The acyclic logic program $L_M$ is given by the following clauses:

$$CS = 1 \Leftarrow U_s = 1; \quad B_1 = 1 \Leftarrow CS = 1;$$
$$B_2 = 1 \Leftarrow CS = 1; \quad S = 1 \Leftarrow B_1 = 1 \vee B_2 = 1.$$

Finally, the probability function $P_M$ on $C_M$'s total choices $c_1 = \{U_s = 1\}$ and $c_2 = \{U_s = 0\}$ is defined by $P_M(c_1) = P(U_s = 1)$ and $P_M(c_2) = P(U_s = 0)$. □

## 5 WEAK AND ACTUAL CAUSES

In this section, we extend the notions of weak and actual cause by Halpern and Pearl [17] to ICL-theories. Informally, the main idea behind this extension is to define weak and actual causes in ICL-theories $T$ as weak and actual causes in their associated causal models $M_T$.

Observe that even though [17] defines the notions of weak and actual cause only for the restricted case of a finite number of endogenous variables, the extended version of [17] also describes how these notions can be generalized to the infinite case. It also gives an example, which deals with infinite weak and actual causes, where such a generalization is necessary. In the sequel, we consider only the original definitions from [17], and we disallow infinite weak and actual causes, to avoid the above-mentioned problems.

We first recall weak and actual causes from [17]. A *primitive event* is an expression $Y = y$, where $Y$ is a variable and $y$ is a value for $Y$. The set of *events* is the closure of the set of primitive events under the Boolean operators $\neg$ and $\wedge$. The *truth* of an event $\phi$ in $M = (U, V, F)$ under $u \in D(U)$, denoted $(M, u) \models \phi$, is inductively defined by:

- $(M, u) \models Y = y$ iff $Y_M(u) = y$;
- $(M, u) \models \neg \phi$ iff $(M, u) \models \phi$ does not hold;
- $(M, u) \models \phi \wedge \psi$ iff $(M, u) \models \phi$ and $(M, u) \models \psi$.

We use $\phi(u)$ to abbreviate $(M, u) \models \phi$. For $X \subsetneq V$ and $x \in D(X)$, we use $\phi_x(u)$ to abbreviate $(M_x, u) \models \phi$. For $X = \{X_1, \ldots, X_k\} \subseteq V$ with $k \geq 1$ and $x_i \in D(X_i)$, we use $X = x_1 \cdots x_k$ to abbreviate $X_1 = x_1 \wedge \ldots \wedge X_k = x_k$.

Let $M = (U, V, F)$ be a causal model. Let $X \subseteq V$ and $x \in D(X)$, and let $\phi$ be an event. Then, $X = x$ is a *weak cause* of $\phi$ under $u$ iff the following conditions hold:

**AC1.** $X(u) = x$ and $\phi(u)$.

**AC2.** Some set of variables $W \subseteq V \backslash X$ and some values $\bar{x} \in D(X)$, $w \in D(W)$ exist such that (a) $\neg \phi_{\bar{x}w}(u)$, and (b) $\phi_{xw\hat{z}}(u)$ for all $\hat{Z} \subseteq V \setminus (X \cup W)$ and $\hat{z} = \hat{Z}(u)$.

Moreover, $X = x$ is an *actual cause* of $\phi$ under $u$ iff additionally the following condition is satisfied:

**AC3.** $X$ is minimal. That is, no proper subset of $X$ satisfies both AC1 and AC2.

We are now ready to define the notions of weak and actual cause for ICL-theories as follows.

**Definition 5.1** Let $T = (C, L)$ be an ICL-theory, $\psi$ be a conjunction of atoms, and $\phi$ be a formula. Let $B$ be a total choice for $C$, and $E$ be an execution set. Then, $\psi$ is a *weak* (resp., an *actual*) *cause* of $\phi$ under $B$ and $E$ in $T$ iff $\rho(\psi\theta)$ is a *weak* (resp., an *actual*) *cause* of $\rho(\phi\theta)$ under $u_B$ in $(M_T)_E$ for each ground substitution $\theta$, where $\rho(\delta)$ is obtained from $\delta$ by replacing every ground atom $p$ by $p = \top$.

**Example 5.1** *(mobile robot (continued))* Consider again the mobile robot scenario described in Example 3.1. Suppose now that there are two objects $o_1$ and $o_2$, and that the robot $r_1$ cannot hold them both in the same time. Consider the ICL-theory $T = (C, L)$, where the choice space is given by $C = \{\{fa_{o,t} = fa(pickUp(o), t), su_{o,t} = su(pickUp(o), t)\} \mid o \in \{o_1, o_2\}, t \in \{0, 1, 2\}\}$, and $L$ is given as in Example 3.1 except that the first clause is replaced by the following two clauses:

$carrying(O, T+1) \Leftarrow at(r_1, Pos, T) \wedge at(O, Pos, T) \wedge$
　　$do(pickUp(O), T) \wedge su(pickUp(O), T) \wedge$
　　$\neg carryingObj(T);$

$carryingObj(T) \Leftarrow carrying(O, T).$

Assume that executing a pickup succeeds at every time point $t \in \{0, 1, 2\}$, which is represented by the total choice $B = \{su_{o,t} \mid o \in \{o_1, o_2\}, t \in \{0, 1, 2\}\}$, and that the robot $r_1$ executes a pickup of $o_1$ at time point 0, a move to $p_1$ at time point 1, and a pickup of $o_2$ at time point 2, which is expressed by the action execution set $E = \{do(pickUp(o_1), 0), do(moveTo(p_1), 1), do(pickUp(o_2), 2)\}$. We now show that $o_1$ being at position $p_2$ at time point 0 is an actual cause of the robot not carrying $o_2$ at time point 2, that is, that $\psi = at(o_1, p_2, 0)$ is an actual cause of $\phi = \neg carrying(o_2, 2)$ under $B$ and $E$ in $T$.

We show that $at(o_1, p_2, 0) = \top$ is an actual cause of $carrying(o_2, 2) = \bot$ under $u_B$ in $(M_T)_E = (U, V, F)$. Obviously, $at(o_1, p_2, 0)(u_B) = \top$ and $carrying(o_2, 2)(u_B) = \bot$ in $(M_T)_E$. That is, (AC1) holds. Consider next $X = \{at(o_1, p_2, 0)\}$, $x = \{(at(o_1, p_2, 0), \top)\}$, $\bar{x} = \{(at(o_1, p_2, 0), \bot)\}$, $W = \{at(o_2, p_1, t) \mid t \in \{0, 1, 2, \ldots\}\}$, and $w = \{(W_i, \top) \mid W_i \in W\}$. We then obtain $carrying(o_2, 2)_{\bar{x}w}(u_B) = \top$. That is, (AC2) (a) holds. Furthermore, we obtain $carrying(o_2, 2)_{xw}(u_B) = \bot$ and $Z_{xw}(u_B) = Z(u_B)$ for $Z = V \backslash (X \cup W)$ (setting $W = w$ and thus inverting $at(o_2, p_1, t)$ for $t \in \{0, 1, 2, \ldots\}$ does not affect other fluents). That is, also (AC2) (b) holds. This shows that $at(o_1, p_2, 0) = \top$ is a weak cause of $carrying(o_2, 2) = \bot$ under $u_B$ in $(M_T)_E$. Since $X$ is a singleton, also (AC3) holds. It thus follows that $at(o_1, p_2, 0) = \top$ is also an actual cause of $carrying(o_2, 2) = \bot$ under $u_B$ in $(M_T)_E$. $\square$

The next example shows that we can also refer to the actually executed actions in weak and actual causes, if we make use of representation (1) for action execution sets.

**Example 5.2** *(waiting collector)* We consider a simplified version of Example 3.1, where (i) we have only the fluent $carrying(T)$ with $T \in \{0, 1, \ldots\}$ and the two actions $wait$ and $pickUp$, and (ii) we assume that the action $pickUp$ can fail or succeed (which depends on the time point of the execution of $pickUp$). Consider the ICL-theory $T = (C, L)$, where the choice space is given by $C = \{\{su(pickUp, t), fa(pickUp, t)\} \mid t \in \{0, 1\}\}$, and $L$ describes the dynamics of this simple scenario by means of the following two clauses:

$carrying(T+1) \Leftarrow do(pickUp, T) \wedge su(pickUp, T) \wedge$
　　$\neg do(wait, T);$

$carrying(T+1) \Leftarrow carrying(T).$

Assume that executing a pickup succeeds at time point 0, but fails at time point 1, which is represented by the total choice $B = \{fa(pickUp, 1), su(pickUp, 0)\}$, and that the robot $r_1$ waits at time point 0 and executes a pickup at time point 1, which is expressed by the action execution set $E' = \{do(wait, 0), do(pickUp, 1)\}$. It can now be shown [10] that the robot's waiting at time point 0 is an actual cause of not carrying the object $o_1$ at time point 2, that is, that $\psi = do(wait, 0)$ is an actual cause of $\phi = \neg carrying(2)$ under $B$ and $E = \emptyset$ in $T' = (C, L')$, where $L' = L \cup \{do(wait, 0) \Leftarrow \top, do(pickUp, 1) \Leftarrow \top\}$. $\square$

## 6 EXPLANATIONS

In this section, we extend the concept of a (causal) explanation by Halpern and Pearl [18] to ICL-theories.

We first recall the concept of an explanation from [18]. Let $M = (U, V, F)$ be a causal model. Let $X \subseteq V$ and $x \in D(X)$, $\phi$ be an event, and $\mathcal{C} \subseteq D(U)$ be a set of contexts. Then, $X = x$ is an *explanation* of $\phi$ relative to $\mathcal{C}$ iff the following conditions (EX1)–(EX4) hold:

**EX1.** $\phi(u)$ holds, for each context $u \in \mathcal{C}$.

**EX2.** $X = x$ is a weak cause of $\phi$ under every $u \in \mathcal{C}$ such that $X(u) = x$.

**EX3.** $X$ is minimal. That is, for every $X' \subset X$, some $u \in \mathcal{C}$ exists such that $X'(u) = x|X'$ and $X' = x|X'$ is not a weak cause of $\phi$ under $u$.

**EX4.** $X(u) = x$ and $X(u') \neq x$ for some $u, u' \in \mathcal{C}$.

We now define explanations for ICL-theories as follows.

**Definition 6.1** Let $T = (C, L)$ be an ICL-theory, $\psi$ be a conjunction of atoms, and $\phi$ be a formula. Let $\mathcal{B}$ be a set of total choices for $C$, and $E$ be an execution set. Then, $\psi$ is an *explanation* of $\phi$ under $\mathcal{B}$ and $E$ in $T$ iff $\rho(\psi\theta)$ is an explanation of $\rho(\phi\theta)$ under $\{u_B \mid B \in \mathcal{B}\}$ in $(M_T)_E$ for each ground substitution $\theta$, where $\rho$ is defined as in Def. 5.1.

**Example 6.1** *(mobile robot (continued))* Consider a new version of the mobile robot scenario in Example 3.1, where we have the new fluent $carryingObj(T)$, $T \in \{0, 1, \ldots\}$, and $carryingObj(t)$ means that the robot $r_1$ is carrying an object at time point $t$. Let the ICL-theory $T = (L, C)$ be defined by the choice space $C = \{\{su(pickUp(o_1), 0), fa(pickUp(o_1), 0)\}\}$ and the logic program $L$ as in Example 3.1 except that the last two clauses are replaced by:

$$carryingObj(T) \Leftarrow carrying(O, T);$$
$$at(o_1, p_1, 0) \Leftarrow \top; \quad at(r_1, p_1, 0) \Leftarrow \top;$$
$$su(pickUp(o_1), 1) \Leftarrow \top.$$

Assume that executing a pickup of $o_1$ either succeeds or fails at time point 0, which is expressed by the set of total choices $\mathcal{B} = \{B_1 = \{su(pickUp(o_1), 0)\}, B_2 = \{fa(pickUp(o_1), 0)\}\}$, and that the robot $r_1$ executes a pickup of $o_1$ at time points 0 and 1, which is represented by the action execution set $E = \{do(pickUp(o_1), 0), do(pickUp(o_1), 1)\}$. It can now be shown [10] that carrying $o_1$ at time point 1 is an explanation of carrying an object at time point 2, that is, $\psi = carrying(o_1, 1)$ is an explanation of $\phi = carryingObj(2)$ under $\mathcal{B}$ and $E$ in $T$. □

## 7 PARTIAL EXPLANATIONS

We finally extend the notions of partial and $\alpha$-partial explanations by Halpern and Pearl [18] to PICL-theories.

We first recall the notions of partial and $\alpha$-partial explanations and of explanatory power from [18]. Let $M = (U, V, F)$ be a causal model. Let $X \subseteq V$ and $x \in D(X)$, let $\phi$ be an event, let $\mathcal{C} \subseteq D(U)$ such that $\phi(u)$ for all $u \in \mathcal{C}$. The expression $\mathcal{C}^\phi_{X=x}$ denotes the unique largest subset $\mathcal{C}'$ of $\mathcal{C}$ such that $X = x$ is an explanation of $\phi$ relative to $\mathcal{C}'$.

**Proposition 7.1 (See [8])** *If $X = x$ is an explanation of $\phi$ relative to some $\mathcal{C}' \subseteq \mathcal{C}$, then $\mathcal{C}^\phi_{X=x}$ is defined, and it contains all $u \in \mathcal{C}$ such that either $X(u) \neq x$, or $X(u) = x$ and $X = x$ is a weak cause of $\phi$ under $u$.*

Let $P$ be a probability function on $\mathcal{C}$, and define

$$P(\mathcal{C}^\phi_{X=x} \mid X = x) \;=\; \sum_{\substack{u \in \mathcal{C}^\phi_{X=x} \\ X(u) = x}} P(u) \;/\; \sum_{\substack{u \in \mathcal{C} \\ X(u) = x}} P(u).$$

Then, $X = x$ is called an $\alpha$-*partial explanation* of $\phi$ relative to $(\mathcal{C}, P)$ iff $\mathcal{C}^\phi_{X=x}$ is defined and $P(\mathcal{C}^\phi_{X=x} \mid X=x) \geq \alpha$. Moreover, $X = x$ is a *partial explanation* of $\phi$ relative to $(\mathcal{C}, P)$ iff $X = x$ is an $\alpha$-partial explanation of $\phi$ relative to $(\mathcal{C}, P)$ for some $\alpha > 0$. Then, $P(\mathcal{C}^\phi_{X=x} \mid X = x)$ is called its *explanatory power* (or *goodness*).

We are now ready to define $\alpha$-partial explanations for PICL-theories. This then implicitly also defines partial explanations and their explanatory power for PICL-theories.

**Definition 7.1** Let $T = ((C, L), P)$ be a PICL-theory, let $\psi$ be a conjunction of atoms, and let $\phi$ be a formula. Let $\mathcal{B}$ be a set of total choices for $C$, and let $E$ be an action execution set. Then, $\psi$ is an $\alpha$-*partial explanation* of $\phi$ under $\mathcal{B}$ and $E$ in $T$ iff $\rho(\psi\theta)$ is an $\alpha$-partial explanation of $\rho(\phi\theta)$ relative to $(\{u_B \mid B \in \mathcal{B}\}, P_T)$ in $(M_T)_E$ for every ground substitution $\theta$, where $\rho$ is defined as in Def. 5.1.

**Example 7.1** *(mobile robot (continued))* We consider another version of the mobile robot scenario in Example 3.1, where we assume two positions $p_1$ and $p_2$, and two objects $o_1$ and $o_2$, which the robot $r_1$ cannot hold in the same time. Let the PICL-theory $T = ((L, C), P)$ be given by the choice space $C = \{\{su_{\bullet,t} = su(pickUp(o), t), fa_{o,t} = fa(pickUp(o), t)\} \mid (o, t) \in \{(o_1, 0), (o_2, 2)\}\}$, the logic program $L$ as in Example 3.1 except that the last two clauses are replaced by the following clauses:

$$at(o_1, p_1, 0) \Leftarrow \top; \; at(o_2, p_2, 0) \Leftarrow \top; \; at(r_1, p_1, 0) \Leftarrow \top,$$

and the probability function $P$ obtained from $P(fa_{o,t}) = 0.3$ and $P(su_{o,t}) = 0.7$ by assuming independence between the $A \in C$. Suppose that either (a) executing a pickup of $o_1$ and $o_2$ succeeds at time points 0 and 2, respectively, or (b) executing a pickup of $o_2$ fails at time point 2, expressed by the set of total choices $\mathcal{B} = \{B_1, B_2, B_3\}$, where

$$B_1 = \{su(pickUp(o_1), 0), su(pickUp(o_2), 2)\},$$
$$B_2 = \{su(pickUp(o_1), 0), fa(pickUp(o_2), 2)\},$$
$$B_3 = \{fa(pickUp(o_1), 0), fa(pickUp(o_2), 2)\}.$$

Furthermore, assume that the robot $r_1$ executes a pickup of the object $o_1$ at time point 0, a move to $p_1$ at time point 1, and a pickup of $o_2$ at time point 2, which is expressed by the action execution set $E = \{do(pickUp(o_1), 0), do(moveTo(p_1), 1), do(pickUp(o_2), 2)\}$. It can now be shown [10] that carrying $o_1$ at time point 1 is an $\alpha$-partial explanation of not carrying $o_2$ at time point 3, that is, that $\psi = carrying(o_1, 1)$ is an $\alpha$-partial explanation of $\phi = \neg carrying(o_2, 3)$ under $\mathcal{B}$ and $E$ in $T$, where $\alpha = P(B_1) / (P(B_1) + P(B_2)) = 0.49 / (0.49 + 0.21) = 0.7$. □

## 8  SUMMARY AND OUTLOOK

We presented a combination of Pearl's structural-model approach to causality with Poole's independent choice logic. We showed how probabilistic theories in the independent choice logic can be mapped to probabilistic causal models. This mapping provides the independent choice logic with appealing concepts of causality and explanation from the structural-model approach. We illustrated this along Halpern and Pearl's sophisticated notions of actual cause, explanation, and partial explanation. Moreover, this mapping also adds first-order modeling capabilities and explicit actions to the structural-model approach.

An interesting topic of future research is to explore the counterparts of other important concepts from the structural-model approach (such as probabilistic counterfactuals and probabilistic causal independence) in the independent choice logic. Furthermore, it would be interesting to explore a generalization of the presented approach to non-acyclic logic programs, which may then involve non-recursive causal models. Finally, another interesting topic is to explore how to define the concepts from the structural-model approach directly in the independent choice logic.

**Acknowledgments**

This work was partially supported by a Marie Curie Individual Fellowship of the European Community programme "Human Potential" under contract number HPMF-CT-2001-001286 (Disclaimer: The authors are solely responsible for information communicated and the European Commission is not responsible for any views or results expressed) and by the Austrian Science Fund under project Z29-N04. We thank the reviewers for their constructive comments, which helped to improve our work.

**References**

[1] F. Bacchus, J. Y. Halpern, and H. J. Levesque. Reasoning about noisy sensors and effectors in the situation calculus. *Artif. Intell.*, 111(1-2):171–208, 1999.

[2] A. Balke and J. Pearl. Probabilistic evaluation of counterfactual queries. In *Proc. AAAI-94*, pp. 230–237, 1994.

[3] C. Baral, N. Tran, and L.-C. Tuan. Reasoning about actions in a probabilistic setting. In *Proc. AAAI-02*, pp. 507–512.

[4] C. Boutilier, R. Reiter, and B. Price. Symbolic dynamic programming for first-order MDPs. In *Proc. IJCAI-01*, pp. 690–700, 2001.

[5] C. Boutilier, R. Reiter, M. Soutchanski, and S. Thrun. Decision-theoretic, high-level agent programming in the situation calculus. In *Proc. AAAI-00*, pp. 355–362, 2000.

[6] U. Chajewska and J. Y. Halpern. Defining explanation in probabilistic systems. In *Proc. UAI-97*, pp. 62–71, 1997.

[7] T. Eiter and T. Lukasiewicz. Complexity results for structure-based causality. In *Proc. IJCAI-01*, pp. 35–40, 2001. Extended version: *Artif. Intell.*, 142(1), 53–89, 2002.

[8] T. Eiter and T. Lukasiewicz. Complexity results for explanations in the structural-model approach. In *Proc. KR-02*, pp. 49–60, 2002. Extended version: *Artif. Intell.*, to appear.

[9] T. Eiter and T. Lukasiewicz. Causes and explanations in the structural-model approach: Tractable cases. In *Proc. UAI-02*, pp. 146–153, 2002.

[10] A. Finzi and T. Lukasiewicz. Structure-based causes and explanations in the independent choice logic. Technical Report INFSYS RR-1843-03-06, Institut für Informationssysteme, TU Wien, April 2003.

[11] A. Finzi and F. Pirri. Combining probabilities, failures and safety in robot control. In *Proc. IJCAI-01*, pp. 1331–1336.

[12] D. Galles and J. Pearl. Axioms of causal relevance. *Artif. Intell.*, 97:9–43, 1997.

[13] P. Gärdenfors. *Knowledge in Flux*. MIT Press, 1988.

[14] H. Geffner. Causal theories for nonmonotonic reasoning. In *Proc. AAAI-90*, pp. 524–530, 1990.

[15] H. Geffner. *Default Reasoning: Causal and Conditional Theories*. MIT Press, 1992.

[16] J. Y. Halpern. Axiomatizing causal reasoning. *J. Artif. Intell. Res.*, 12:317–337, 2000.

[17] J. Y. Halpern and J. Pearl. Causes and explanations: A structural-model approach – Part I: Causes. In *Proc. UAI-01*, pp. 194–202, 2001.

[18] J. Y. Halpern and J. Pearl. Causes and explanations: A structural-model approach – Part II: Explanations. In *Proc. IJCAI-01*, pp. 27–34, 2001.

[19] C. G. Hempel. *Aspects of Scientific Explanation*. Free Press, 1965.

[20] M. Henrion and M. J. Druzdzel. Qualitative propagation and scenario-based approaches to explanation of probabilistic reasoning. In *Uncertainty in Artificial Intelligence 6*, pp. 17–32. Elsevier Science, 1990.

[21] M. Hopkins. Strategies for determining causes of reported events. In *Proc. AAAI-02*, pp. 546–552, 2002.

[22] M. Hopkins and J. Pearl. Causality and counterfactuals in the situation calculus. Technical Report R-301, UCLA Cognitive Systems Lab, 2002.

[23] M. Hopkins and J. Pearl. Clarifying the usage of structural models for commonsense causal reasoning. In *Proc. of the AAAI Spring Symposium on Logical Formalizations of Commonsense Reasoning*, 2003.

[24] V. Lifschitz. On the logic of causal explanation. *Artif. Intell.*, 96:451–465, 1997.

[25] P. Mateus, A. Pacheco, and J. Pinto. Observations and the probabilistic situation calculus. In *Proc. KR-02*, pp. 327–338, 2002.

[26] N. McCain and H. Turner. Causal theories of action and change. In *Proc. AAAI-97*, pp. 460–465, 1997.

[27] J. Pearl. *Probabilistic Reasoning in Intelligent Systems: Networks of Plausible Inference*. Morgan Kaufmann, 1988.

[28] J. Pearl. Reasoning with cause and effect. In *Proc. IJCAI-99*, pp. 1437–1449, 1999.

[29] J. Pearl. *Causality: Models, Reasoning, and Inference*. Cambridge University Press, 2000.

[30] D. Poole. The independent choice logic for modelling multiple agents under uncertainty. *Artif. Intell.*, 94:7–56, 1997.

[31] D. Poole. Decision theory, the situation calculus and conditional plans. *Electronic Transactions on Artificial Intelligence*, 2(1-2):105–158, 1998.

[32] D. Poole. Logic, knowledge representation, and Bayesian decision theory. In *Proc. CL-00*, pp. 70–86, 2000.

[33] R. Reiter. *Knowledge in Action: Logical Foundations for Specifying and Implementing Dynamical Systems*. 2001.

[34] W. C. Salmon. *Four Decades of Scientific Explanation*. University of Minnesota Press, 1989.

[35] S. E. Shimony. Explanation, irrelevance, and statistical independence. In *Proc. AAAI-91*, pp. 482–487, 1991.

[36] H. Turner. A logic of universal causation. *Artif. Intell.*, 113:87–123, 1999.